\documentclass[runningheads]{llncs}
\usepackage[T1]{fontenc}
\usepackage{graphicx}
\usepackage{bbm}
\usepackage{amssymb}
\usepackage{amsfonts}
\usepackage{amsmath}
\usepackage{stfloats}
\usepackage{multirow}
\usepackage{array}
\usepackage{booktabs}
\usepackage{makecell}
\usepackage{color}

\usepackage[misc]{ifsym}
\usepackage{lipsum}

\begin{document}
\title{Prompt-based Grouping Transformer for Nucleus Detection and Classification}
\titlerunning{Prompt-based Grouping Transformer}
%
%

\author{
Junjia Huang\inst{1,2}$^{*}$
\and 
Haofeng Li\inst{2,3}$^{*}$
\and 
Weijun Sun\inst{4}
\and 
Xiang Wan\inst{2}
\and 
Guanbin Li\inst{1}$^{(\textrm{\Letter})}$
}
%
\authorrunning{J. Huang et al.}
%

\institute{School of Computer Science and Engineering, Research Institute of Sun Yat-sen University in Shenzhen, Sun Yat-sen University, Guangzhou, China
\and
Shenzhen Research Institute of Big Data, Shenzhen, China 
\and 
The Chinese University of Hong Kong, Shenzhen, China
\and
Guangdong University of Technology, Guangzhou, China\\ 
\email{liguanbin@mail.sysu.edu.cn}
}

\maketitle              

\newcommand\blfootnote[1]{%
\begingroup
\renewcommand\thefootnote{}\footnote{#1}%
\addtocounter{footnote}{-4}%
\endgroup
}

\begin{abstract}
Automatic nuclei detection and classification can produce effective information for disease diagnosis. Most existing methods classify nuclei independently or do not make full use of the semantic similarity between nuclei and their grouping features. In this paper, we propose a novel end-to-end nuclei detection and classification framework based on a grouping transformer-based classifier. The nuclei classifier learns and updates the representations of nuclei groups and categories via hierarchically grouping the nucleus embeddings. Then the cell types are predicted with the pairwise correlations between categorical embeddings and nucleus features. For the efficiency of the fully transformer-based framework, we take the nucleus group embeddings as the input prompts of backbone, which helps harvest grouping guided features by tuning only the prompts instead of the whole backbone. Experimental results show that the proposed method significantly outperforms the existing models on three datasets.  
\blfootnote{This work was supported in part by the Chinese Key-Area Research and Development Program of Guangdong Province (2020B0101350001), 
in part by the National Natural Science Foundation of China (No.~62102267, NO.~61976250), in part by the Guangdong Basic and Applied Basic Research Foundation (2023A1515011464, 2020B1515020048), in part by the Shenzhen Science and Technology Program (JCYJ20220818103001002, JCYJ20220530141211024), and the Guangdong Provincial Key Laboratory of Big Data Computing, The Chinese University of Hong Kong, Shenzhen. 
Guanbin Li is the corresponding author.  
Junjia Huang and Haofeng Li contribute equally to this work.
}
\end{abstract}
\section{Introduction}

Nucleus classification is to identify the cell types from digital pathology image, 
assisting pathologists in cancer diagnosis and prognosis~\cite{aeffner2019introduction,sirinukunwattana2016locality}. For example, the involvement of tumor-infiltrating lymphocytes (TILs) is a critical prognostic variable for the evaluation of breast/lung cancer~\cite{salgado2015evaluation,bremnes2016role}. 
It is a challenge to infer the nucleus types due to the diversity and unbalanced distribution of nuclei. 
Thus, we aim to automatically classify cell nuclei in pathological images.

A number of methods~\cite{zeng2019ric,graham2019hover,liu2020scam,zhou2021ssmd,doan2022sonnet,kiran2022denseres,lou2022pixel,lou2023multi} have been proposed for automatic nuclei segmentation and classification. Most of them use a U-shape model~\cite{ronneberger2015u} for training to produce dense predictions with expensive pixel-level labels. 
In this paper, we aim to obtain the location and category of cells, which only needs affordable labels of centroids or bounding boxes. 
The task can be solved by generic object detector~\cite{li2019detection,nair2021mitotic,obeid2022nucdetr}, but they are usually built for everyday objects whose positions and combinations are quite random. 
Differently, in pathological images, experts often identify nuclear communities via their relationships and spatial distribution. Some recent methods resort to the spatial contexts among nuclei. Abousamra \textit{et al.}~\cite{abousamra2021multi} adopt a spatial statistical function to model the local density of cells. Hassan \textit{et al.}~\cite{hassan2022nucleus} build a location-based graph for nuclei classification. 
However, the semantics similarity and dissimilarity between nucleus instances as well as the category representations have not been fully exploited.

Based on these observations, we develop a learnable Grouping Transformer based Classifier (GTC) that leverages the similarity between nuclei and their cluster representations to infer their types. 
Specifically, we define a number of nucleus clusters with learnable initial embeddings, and assign nucleus instances to their most correlated clusters by computing the correlations between clusters and nuclei. 
Next, the cluster embeddings are updated with their affiliated instances, and are further grouped into the categorical representations. 
Then, the cell types can be well estimated using the correlations between the nuclei and the categorical embeddings. 
We propose a novel fully transformer-based framework for nuclei detection and classification, by integrating a backbone, a centroid detector, and the grouping-based classifier. 
However, the transformer framework has a relatively large number of parameters, which could cause high costs in fine-tuning the whole model on large datasets. 
On the other hand, there exist domain gaps in the pathological images of different organs, staining, and institutions, which makes it necessary to fine-tune models to new applications. Thus, it is of great significance to tune our proposed transformer framework efficiently.

Inspired by the prompt tuning methods~\cite{lester-etal-2021-power,liu-etal-2022-p,jia2022visual} which train continuous prompts with frozen pretrained models for natural language processing tasks, we propose a grouping prompt based learning strategy for efficient tuning. We prepend the embeddings of nucleus clusters to the input space and freeze the entire pre-trained transformer backbone so that these group embeddings act as prompt information to help the backbone extract grouping-aware features. 
Our contributions are: (1) a prompt-based grouping transformer framework for end-to-end detection and classification of nuclei; (2) a novel grouping prompt learning mechanism that exploits nucleus clusters to guide feature learning with low tuning costs; (3) Experimental results show that our method achieves the state-of-the-art on three public benchmarks.

\section{Methodology}
\begin{figure}[!t]
\centering
\includegraphics[width=0.9\textwidth]{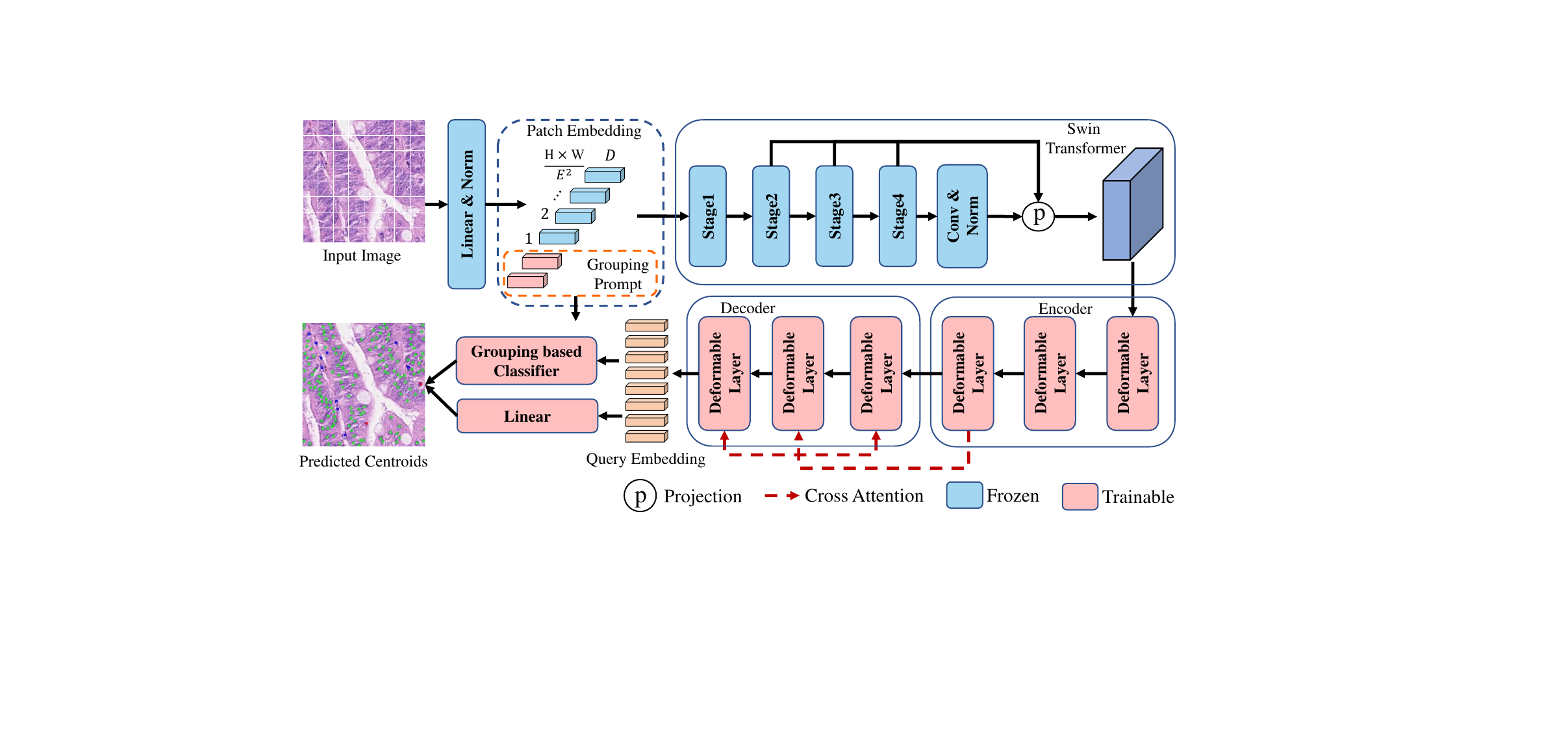}
\caption{The architecture of Prompt-based Grouping Transformer.}
\label{framework}
\end{figure}
As shown in Fig.~\ref{framework}, We propose a novel framework, Prompt-based Grouping Transformer (PGT), which directly outputs the coordinates of nuclei centroids and leverages \textit{grouping prompts} for cell-type prediction. In the architecture, the detection and classification parts are interdependent and can be trained together. The proposed framework consists of a transformer-based nucleus detector, a grouping transformer-based classifier, and a grouping prompt learning strategy, which are presented in the following.

\subsection{Transformer-based Centroid Detector}
\textbf{Backbone.} We adopt Swin Transformer~\cite{liu2021swin} as the backbone to learn deep features. The pixel-level feature maps output from Stage 2 to Stage 4 of the backbone are extracted. Then the Stage-4 feature map is downsampled with a $3\times 3$ convolution of stride 2 to yield another lower-resolution feature map. We obtain four feature maps in total. The channel number of each feature map is aligned via a $1\times 1$ convolution layer and a group normalization operator. 

\noindent\textbf{Encoder and Decoder.} The encoder and decoder have 3 deformable attention layers~\cite{zhu2020deformable}, respectively. The multi-scale feature maps output by the backbone are fed into the encoder in which the pixel-level feature vectors in all these feature maps are updated via deformable self-attention. After the attention layers, we send each feature vector into 2 fully connected (FC) layers separately to obtain the fine-grained categorical scores of each pixel. Only the $Q$ feature vectors with the highest confidence are preserved as object embeddings and their position coordinates are recorded as reference points. Each decoder layer utilizes cross-attention to enhance the object embeddings by taking them as queries/values and the updated feature maps as keys. The enhanced query embeddings are fed into 2 FC layers to regress position offsets which are added to and refine the reference points. The reference points output by the last decoder layer are the finally detected nucleus centroids. The last query embeddings from the decoder are sent to the proposed classifier for cell type prediction. 

\subsection{Grouping Transformer based Classifier}
\begin{figure}[!t]
\centering
\includegraphics[width=0.9\textwidth]{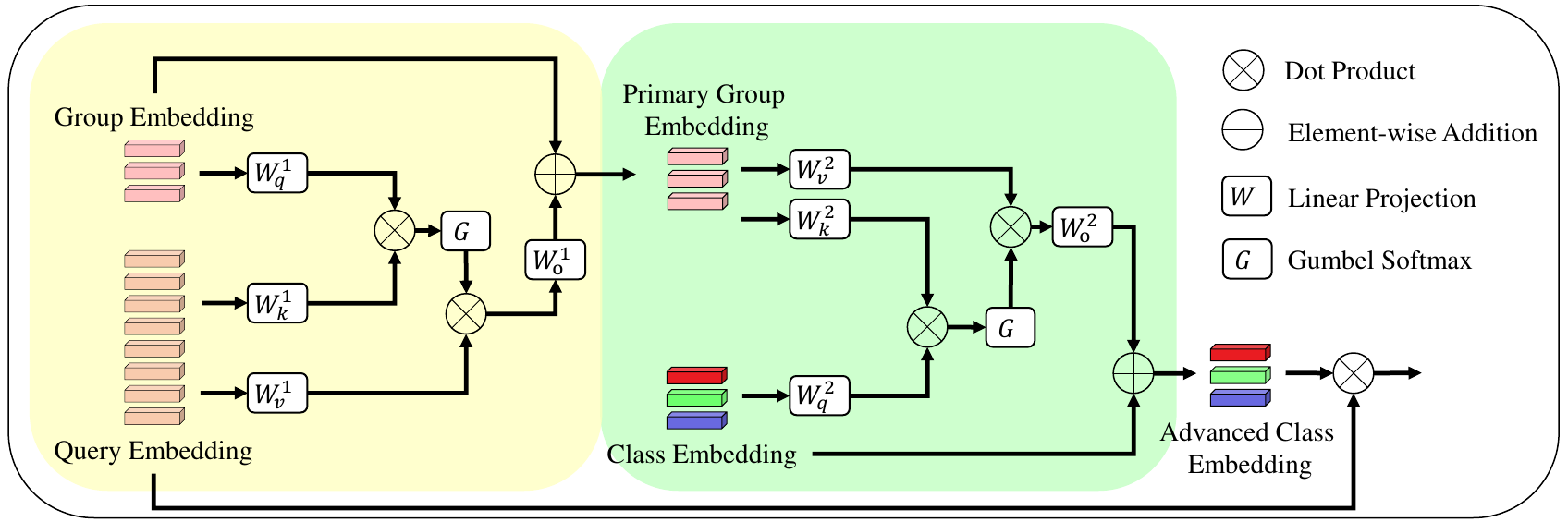}
\caption{The Grouping Transformer based Classifier.} \label{group classifier}
\end{figure}

In Fig.~\ref{group classifier}, we develop a Grouping Transformer based Classifier (GTC) that takes grouping prompts $g \in \mathbb{R}^{G\times D}$ and query embeddings $q \in \mathbb{R}^{Q\times D}$ as inputs, and yields categorical scores for each nucleus query. To divide the queries into primary groups, 
The similarity matrix $S \in \mathbb{R}^{G \times Q}$ between the query embeddings and the grouping prompts is built via inner product and Gumbel-Softmax~\cite{jang2016categorical} operation as Eq.~(\ref{gumbelsoftmax}):
\begin{equation}
    S = \textit{softmax}(W_q^1 g \cdot (W_k^1 q)^T+\gamma/\tau), 
    \label{gumbelsoftmax}
\end{equation} 
where $W_q^1$  and $W_k^1$ are the weights of learnable linear projections, $\gamma \in \mathbb{R}^{G \times Q}$ are i.i.d  random samples drawn from the distribution $Gumbel(0, 1)$ and $\tau$ denotes the Softmax temperature. Then we utilize the hard assignment strategy~\cite{van2017neural,xu2022groupvit} 
and assign the query embedding to different groups as Eq.~(\ref{onehot}):
\begin{equation}
    \hat{S} = \textit{one-hot}(\textit{argmax}(S))+S-sg(S), 
    \label{onehot}
\end{equation}
where $\textit{argmax}(S)$ returns a $1\times Q$ vector, and $\textit{one-hot}(\cdot)$ converts the vector to a binary $G\times Q$ matrix. $sg$ is the stop gradient operator for better training of the one-hot function~\cite{van2017neural,xu2022groupvit}. Then we merge the embeddings belonging to the same group into a primary group via Eq.~(\ref{merge}): 
\begin{equation}
    g_p = g + W_o^1\frac{\hat{S}\cdot W_v^1 q}{\sum^G_{i=1} \hat{S}_i}
    \label{merge}
\end{equation}
where $g_p$ denotes the embeddings of primary groups, $W_v^1$ and $W_o^1$ are learnable linear weights. To separate the primary groups into the cell categories, we measure the similar matrix between the primary groups $g_p$ and learnable class embeddings $c_e \in \mathbb{R}^{C\times D}$ to yield advanced class embeddings $c_a \in \mathbb{R}^{C\times D}$, in the same way as Eq.\eqref{gumbelsoftmax}--\eqref{merge}. 
To classify each centroid query, we measure the similarity between each query embedding and the advanced class embeddings. The category whose advanced embedding is most similar to a query, is assigned to the centroid query. The classification results $c \in \mathbb{R}^{C \times Q}$ are computed as: $c = c_a \cdot q^T$.

\subsection{Loss Function} 
The proposed method outputs a set of centroid proposals $\{(x_q, y_q)|q\in \{1,\cdots, Q\}\}$ with a decoder layer, and their corresponding cell-type scores $\{c_q|q\in \{1,\cdots, Q\}\}$ with our proposed classifier. To compute the loss with detected centroids, we use the Hungarian algorithm~\cite{kuhn1955hungarian} to assign $K$ target centroids (ground truth) 
to proposal centroids and get $P$ positive (matched) samples and $Q-P$ negative (unmatched) samples. The overall loss is defined as Eq.~(\ref{eq:overall_loss}):
\begin{equation}\label{eq:overall_loss}
    L(y, \hat{y}) = \frac{1}{P} \sum^P_{i=1} \left( \omega_1 ||(x_i, y_i)-(\hat{x}_i, \hat{y}_i)||_2^2+\omega_2 \textit{FL}(c_i, \hat{c}_i) \right) +\omega_3 \sum^Q_{j=P+1} \textit{FL}(c_j, \hat{c}_j), 
\end{equation}
where $\omega_1, \omega_2, \omega_3$ are weight terms, $(x_i, y_i)$ is the $i^{th}$ matched centroid coordinates, $(\hat{x}_i, \hat{y}_i)$ is the target coordinates. $c_i$ and $c_j$ denote the categorical scores of matched and unmatched samples, respectively. As the target of unmatched samples, $\hat{c}_j$ is set to an empty category. $\textit{FL}(\cdot)$ is the Focal Loss~\cite{lin2017focal} for training the proposed classifier. 
We adopt the deep supervision strategy~\cite{zhu2020deformable}. 
In the training, each decoder layer produces the side outputs of centroids and query embeddings that are fed into a GTC for classifying nuclei. For the 3 decoder layers, they yield 3 sets of detection and classification results for the loss in Eq.~(\ref{eq:overall_loss}).

\subsection{Grouping Prompts based Tuning}
\begin{figure}[!t]
\centering
\includegraphics[width=0.85\textwidth]{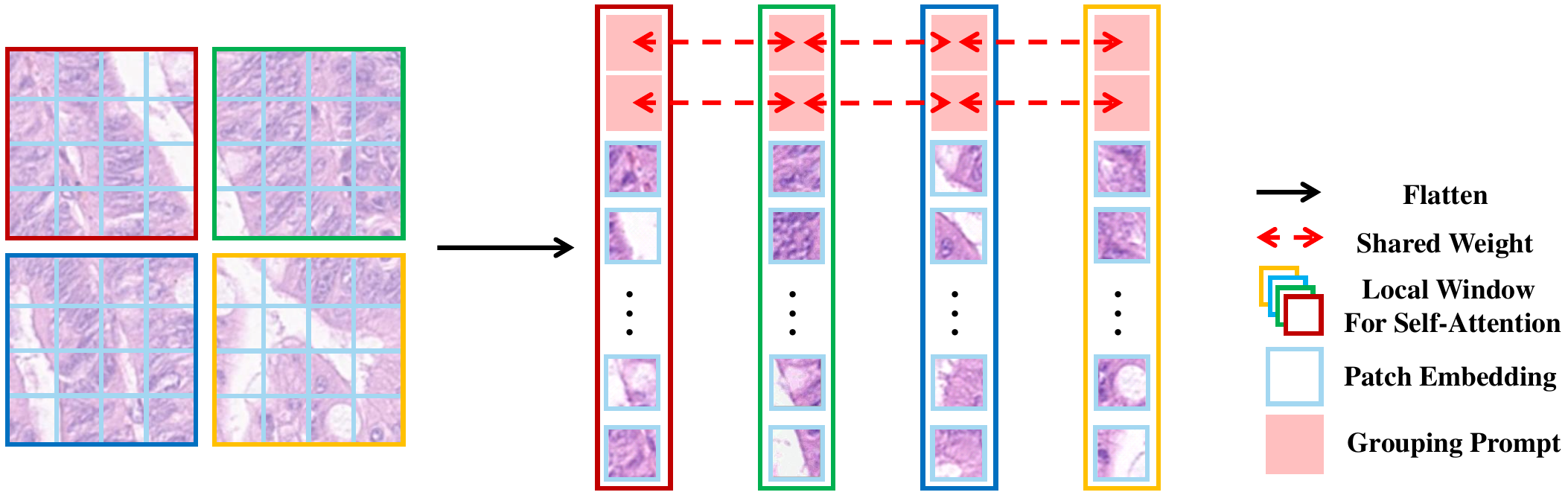}
\caption{The inputs with grouping prompts of the Shift-Window transformer backbone.} \label{swin prompt}
\end{figure}
To avoid the inefficient fine-tuning of the backbone, we propose a new and simple learning strategy based on grouping prompts, as shown in Fig.~\ref{framework}. We inject a set of prompt embeddings as extra input of the Swin-Transformer~\cite{liu2021swin}, and only tune the prompts instead of the backbone. To learn group-aware representations, we further propose to share the embeddings of prompts with those of initial groups in the proposed GTC. Such prompt embeddings are define as \textit{Grouping Prompts}.

For a typical Swin-Transformer backbone, an input pathological image $I\in \mathbb{R}^{H\times W\times 3}$ is divided into $\frac{HW}{E^2}$ image patches of size $E\times E$. We first embed each image patch into a \textit{D}-dimensional latent space via a linear projection. Then we randomly initialize the grouping prompts $g\in \mathbb{R}^{G\times D}$ as learnable parameters, and concatenate them with the patch embeddings as input. Note that in the backbone, input patch embeddings are separated into different local windows and the grouping prompts are also inserted into each window, as shown in Fig.~\ref{swin prompt}. 
Our proposed grouping prompt based learning consists of two phases, pre-tuning and prompt-tuning. In the pre-tuning phase, we adopt the Swin-b backbone pre-trained on ImageNet, replace the GTC head in our model (Fig.~\ref{framework}) with 2 FC layers, and train the overall framework without prompts and GTC. In the prompt-tuning phase, grouping prompts are added to the input of the backbone and GTC, while the backbone parameters are frozen. 

\section{Experiments and Results}
\subsection{Datasets and Implementation Details}
\noindent\textbf{CoNSeP}\footnote{https://warwick.ac.uk/fac/cross\_fac/tia/data/hovernet/}~\cite{graham2019hover} is a colorectal nuclear dataset with three types, consisting of 41 H\&E stained image tiles from 16 colorectal adenocarcinoma whole-slide images (WSIs). The WSIs are at 20$\times$ magnification and the size of the slides is 500x500. We split them following the official partition~\cite{graham2019hover,abousamra2021multi}.

\noindent\textbf{BRCA-M2C}\footnote{https://github.com/TopoXLab/Dataset-BRCA-M2C/}~\cite{abousamra2021multi} is a breast cancer dataset with three types and consists of 120 image tiles from 113 patients. The WSIs are at 20$\times$ magnification and the size of the slides ranges from $465\times 465$ to $504 \times 504$. We follow the work~\cite{abousamra2021multi} to apply the SLIC~\cite{achanta2012slic} algorithm to generate superpixels as instances and split them into 80/10/30 slides for training/validation/testing. 

\noindent\textbf{Lizard}\footnote{https://warwick.ac.uk/fac/cross\_fac/tia/data/lizard/}~\cite{graham2021lizard} has 291 histology images of colon tissue from six datasets, containing nearly half a million labeled nuclei in H\&E stained colon tissue. The WSIs are at 20$\times$ magnification with an average size of 1,016×917 pixels.

Our implementation and the setting of hyper-parameters are based on MMDetection~\cite{chen2019mmdetection}. The number of grouping prompts $G$ is 64. Random crop, flipping, and scaling are used for data augmentation. Our method is trained with PyTorch on a 48GB GPU (NVIDIA A100) for 12-24 hours (depending on the dataset size). More details are listed in the supplementary material.

\subsection{Comparison with the State-of-the-art}
\begin{figure}[!t]
\centering
\includegraphics[width=1\textwidth]{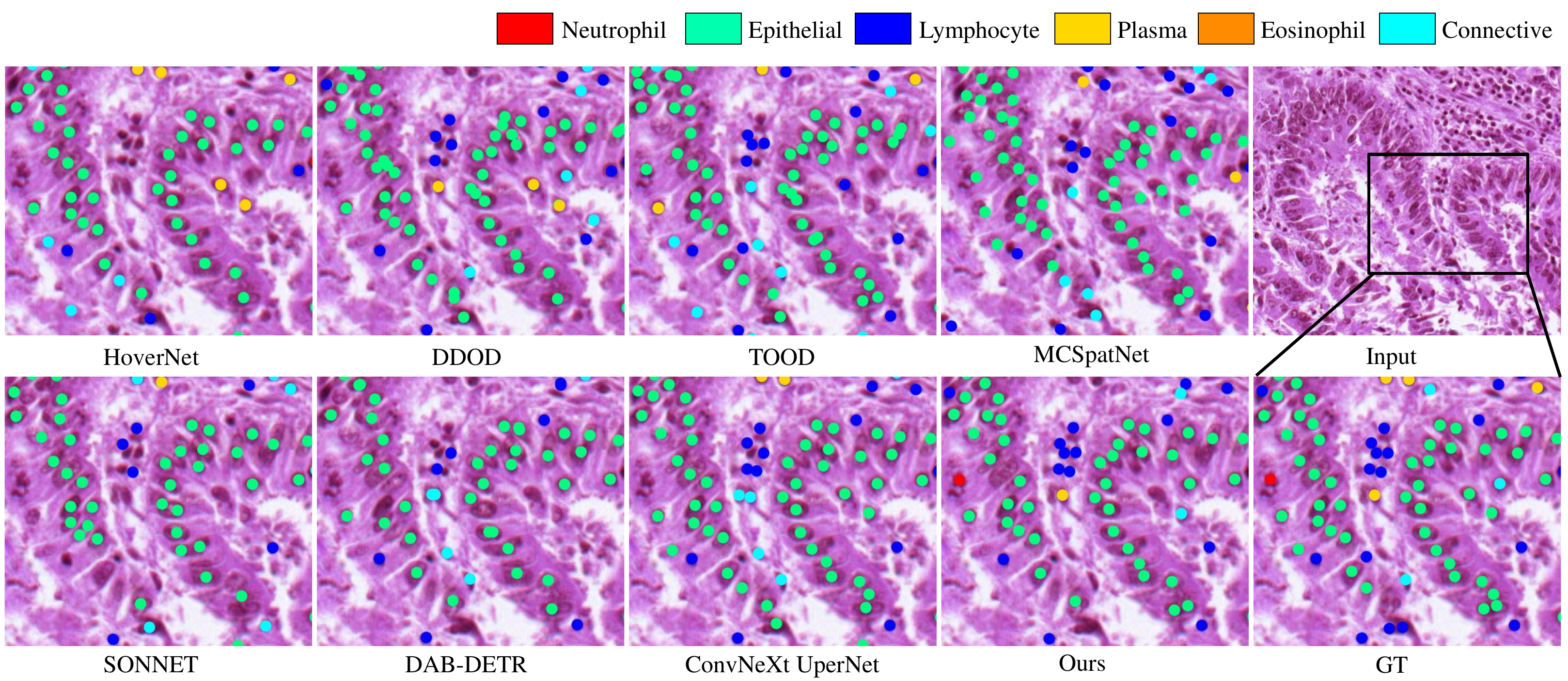}
\caption{The visualization results on CoNSeP dataset.} \label{visual}
\end{figure}

The proposed method is compared with the state-of-the-art models: the existing methods for detecting and classifying cells in pathological images, i.e., HoverNet~\cite{graham2019hover} ,MCSpatNet~\cite{abousamra2021multi}, SONNET~\cite{doan2022sonnet}, and the sate-of-the-art methods for object detection in natural images, i.e., DDOD~\cite{chen2021disentangle}, TOOD~\cite{feng2021tood}, , DAB-DETR~\cite{liu2022dabdetr} and UperNet with ConvNeXt backbone~\cite{liu2022convnet}. As shown in Table \ref{tab:sota}, our method exceeds all the other methods on three benchmarks with both detection and classification metrics. Specifically, on the CoNSeP dataset, our approach achieves 1.6\% higher F-score on the detection ($F_d$) and 1.8\% higher F-score on the classification ($\overline{F_c}$) than the second best methods MCSpatNet~\cite{abousamra2021multi} and UperNet~\cite{liu2022convnet}. On BRCA-M2C dataset, our method has 0.5\% higher $F_d$ and 3.9\% higher $\overline{F_c}$, compared with the second best models MCSpatNet~\cite{abousamra2021multi} and DAB-DETR~\cite{liu2022dabdetr}. Besides, on Lizard dataset, our method outperforms UperNet~\cite{liu2022convnet} by more than 1.5\% and 6.4\% on $F_d$ and $\overline{F_c}$, respectively. Meanwhile, we conduct t-test on CoNSeP dataset for tatistical significance test. The details are listed in the supplementary material.
The visual comparisons are shown in Fig. \ref{visual}. With the context information from surrounding nuclei, our method effectively reduces the misclassification rate of the lymphocytes and neutrophil categories (Blue and Red).

\renewcommand{\arraystretch}{1.2}
\begin{table}[!t]
    \centering
    \caption{Comparison with existing methods on CoNSeP, BRCA-M2C and Lizard. For each dataset, we report the F-score of each class ($F_c^k$), the mean F-score over all classes ($\overline{F_c}$) and the detection F-score ($F_d$). $F_c^{Infl.}$, $F_c^{Epi.}$, $F_c^{Stro.}$, $F_c^{Neu.}$, $F_c^{Lym.}$, $F_c^{Pla.}$, $F_c^{Eos.}$ and $F_c^{Con.}$ denote the F-socre for the inflammatory, epithelial, stromal, neutrophils, lymphocytes, plasma, Eosinophil and connective tissue cells, respectively. For each row, the best result is in \textbf{bold} and the second best is \underline{underlined}.}
    \label{tab:sota}
    \setlength{\tabcolsep}{0.8mm}{
    \scriptsize
    \begin{tabular}{c|l|cccccccc}
    \Xhline{1pt}
         ~ & \multirow{2}*{F-score$\uparrow$} & \multirow{2}*{\makecell[c]{Hovernet \\ \cite{graham2019hover}}} & \multirow{2}*{\makecell[c]{DDOD \\ \cite{chen2021disentangle}}} & \multirow{2}*{\makecell[c]{TOOD \\ \cite{feng2021tood}}} &  \multirow{2}*{\makecell[c]{MCSpatNet \\ \cite{abousamra2021multi}}} & \multirow{2}*{\makecell[c]{SONNET \\ \cite{doan2022sonnet}}} & \multirow{2}*{\makecell[c]{DAB-DETR \\ \cite{liu2022dabdetr}}} &  \multirow{2}*{\makecell[c]{ConvNeXt \\ \cite{liu2022convnet}}} & \multirow{2}*{\makecell[c]{~\\(Ours)}} \\
         ~ & ~ & ~ & ~ & ~ & ~ & ~ &  ~ & ~ & ~ \\
        \hline
        ~ & ~ & 2019 & 2021 & 2021 & 2021 & 2022 & 2022 & 2022 & - \\
         
         \hline
         \hline
         
         \multirow{5}*{\rotatebox{90}{CoNSeP}} & \ $F_c^{Infl.}$ & 0.514 & 0.516 & \underline{0.622} & 0.583 & 0.563 & 0.531 & 0.618 & \textbf{0.623} \\ 
        ~ & \ $F_c^{Epi.}$ & 0.604 & 0.436 & 0.616 & 0.608 & 0.502 & 0.440 & \underline{0.625} & \textbf{0.639} \\ 
        ~ & \ $F_c^{Stro.}$ & 0.391 & 0.429 & 0.382 & 0.527 & 0.366 & 0.443 & \underline{0.542} & \textbf{0.577} \\
        ~ & \ $\overline{F_c}$ & 0.503 & 0.494 & 0.540 & 0.573 & 0.477 & 0.471 & \underline{0.595} & \textbf{0.613} \\ 
        ~ & \ $F_d$ & 0.621 & 0.554 & 0.608 & \underline{0.722} & 0.590 & 0.619 & 0.715 & \textbf{0.738} \\ 
        \hline
        \hline
        \multirow{5}*{\rotatebox{90}{BRCA-M2C}} & \ $F_c^{Infl.}$ & \underline{0.454} & 0.394 & 0.400 & 0.424 & 0.343 & 0.437 & 0.423 & \textbf{0.473} \\ 
        ~ & \ $F_c^{Epi.}$ & 0.577 & 0.544 & 0.559 & 0.627 & 0.411 & 0.634 & \underline{0.636} & \textbf{0.686} \\ 
        ~ & \ $F_c^{Stro.}$ & 0.339 & 0.373 & 0.315 & \underline{0.387} & 0.281 & {0.380} & 0.353 & \textbf{0.409} \\ 
        ~ & \ $\overline{F_c}$ & 0.457 & 0.437 & 0.425 & 0.479 & 0.345 & \underline{0.484} & 0.471 & \textbf{0.523} \\ 
        ~ & \ $F_d$ & 0.74 & 0.659 & 0.662 & \underline{0.794} & 0.653 & 0.705 & 0.785 & \textbf{0.799} \\ 
        \hline
        \hline
        \multirow{7}*{\rotatebox{90}{Lizard}} 
          & \ $F_c^{Neu.}$ & \underline{0.210} & 0.025 & 0.029 & 0.105 & 0.09  & 0.142 & 0.205 & \textbf{0.301} \\ 
        ~ & \ $F_c^{Epi.}$ & 0.665 & 0.584 & 0.615 & 0.601 & 0.599 & 0.653 & \underline{0.714} & \textbf{0.762} \\ 
        ~ & \ $F_c^{Lym.}$ & 0.472 & 0.342 & 0.404 & 0.457 & 0.538 & 0.544 & \underline{0.611} & \textbf{0.664} \\ 
        ~ & \ $F_c^{Pla.}$ & \underline{0.376} & 0.130 & 0.152 & 0.228 & 0.370 & 0.356 & 0.333 & \textbf{0.403} \\ 
        ~ & \ $F_c^{Eos.}$ & 0.367 & 0.124 & 0.157 & 0.220 & 0.365 & 0.295 & \underline{0.403} & \textbf{0.457} \\ 
        ~ & \ $F_c^{Con.}$ & 0.492 & 0.347 & 0.383 & 0.484 & 0.143 & 0.559 & \underline{0.578} & \textbf{0.644} \\ 
        ~ & \ $\overline{F_c}$ & 0.430 & 0.259 & 0.290 & 0.349 & 0.351 & 0.425 & \underline{0.474} & \textbf{0.538} \\ 
        ~ & \ $F_d$ & 0.729 & 0.561 & 0.606 & 0.713 &  0.682 & 0.656 & \underline{0.764} & \textbf{0.779} \\
    \Xhline{1pt}
    \end{tabular}
    }
\end{table}

\renewcommand{\arraystretch}{1.2}
\begin{table}
  \centering
    \caption{Ablation study on CoNSeP. PGT is the overall detection-classification framework. PT denotes training the network with Prompt Tuning. GTC means using the Grouping Transformer-based Classifier. * means freezing the weights of the backbone.}
    \setlength{\tabcolsep}{1.8mm}{
  \scriptsize
  \begin{tabular}{c|cccccc}
    \Xhline{1pt}
    \specialrule{0em}{2pt}{1pt}
    Methods  & $F_c^{Infl.}$ & $F_c^{Epi.}$ & $F_c^{Stro.}$ & $\overline{F_c}$ & $F_d$ & Tuned Params(M) \\
    \midrule
    PGT (Full) & 0.631 & 0.641 & 0.572 & 0.615 & 0.735 & 102.2 \\
    \hline
    w/o GTC \& PT (Baseline)& 0.599 & 0.600 & 0.570 & 0.590 & 0.714 & 95.767 \\ 
    w/o PT* & 0.602 & 0.604 & 0.558 & 0.588 & 0.713 & 15.321 \\
    w/o GTC* & 0.615 & 0.604 & 0.564 & 0.594 & 0.724 & 8.895  \\ 
    w/ detached GTC \& PT* & 0.577 & 0.623 & 0.545 & 0.582 & 0.714 & 15.429 \\
    PGT* (Ours) & \textbf{0.623} & \textbf{0.639} & \textbf{0.577} & \textbf{0.613} & \textbf{0.738} & 15.379 \\
    \specialrule{0em}{1pt}{1pt}
    \Xhline{1pt}
  \end{tabular}
  }
  \label{tab:ab}
\end{table}

\subsection{Ablation Analysis} 
\textbf{The strengths of the grouping transformer based classifier and the grouping prompts} are verified on CoNSeP dataset, as shown in Table \ref{tab:ab}. 
Prompt-based Grouping Transformer (PGT) is our proposed detection and classification architecture with grouping prompts and the GTC (in Fig.~\ref{framework}), while the `Baseline' has no these two settings. 
PT means using naive prompt tuning. GTC means classifying nuclei with the grouping transformer. Our method achieves comparable results to the fully fine-tuning PGT with tuning only 15\% parameters. Compared to the Baseline, our method yields 2.4\% higher $F_d$ and 2.3\% higher $\overline{F_c}$, respectively, which shows the effective combination of the grouping classifier and prompts. 
`detached GTC \& PT' means that group features and prompts are independent. Our method surpasses the detached setting by 2.4\% in $F_d$ and 3.1\% in $\overline{F_c}$, which suggests that sharing embeddings of groups and prompts is effective. 
With a frozen backbone, the performances of `w/o PT' and `w/o GTC' are both dropping, which verifies the strength of the prompt tuning and the GTC module, respectively. 

Table~\ref{tab:number of pge} shows \textbf{the effect of different numbers of grouping prompts} on CoNSeP dataset. When the number of groups is small, the classification result is inferior. When the group number is large than 64, the groups may contain too few nuclei to capture their common patterns. It is suggested to set the group number to a moderate value such as 64. 

\renewcommand{\arraystretch}{1.2}
\begin{table}[!t]
    \centering
    \caption{The effects of the number of grouping prompts $G$ on CoNSeP.}
    \label{tab:number of pge}
    \setlength{\tabcolsep}{1.2mm}{
    \scriptsize
    \begin{tabular}{c|ccccc}
    \Xhline{1pt}
     F-score$\uparrow$ & 8 & 16 & 32 & 64 & 128 \\
    \Xcline{1-6}{0.4pt}
     $F_d$ & 0.727 & 0.724 & 0.726 & \textbf{0.738} & 0.723 \\
     $\overline{F_c}$ & 0.600 & 0.599 & 0.604 & \textbf{0.613} & 0.583\\
    \Xhline{1pt}
    \end{tabular}
    }
\end{table}

\begin{table}[!htb]
\centering
\scriptsize
\caption{ $\overline{F_d}$ denotes the mean of detection F-scores of all testing images. * means p-value$\leq$0.05. ** means p-value$\leq$0.01.}
\label{tab:sst}
\setlength{\tabcolsep}{0.2mm}{
\begin{tabular}{c|cccccccc}
\Xhline{2pt}
F-score$\uparrow$ & \makecell[c]{Hovernet \\ \cite{graham2019hover}
} & \makecell[c]{DDOD\\ \cite{chen2021disentangle}
} & \makecell[c]{TOOD\\ \cite{feng2021tood}
} & \makecell[c]{MCSpatNet \\ \cite{abousamra2021multi}
} & \makecell[c]{SONNET\\ \cite{doan2022sonnet}
} & \makecell[c]{DAT-DETR \\ \cite{liu2022dabdetr}
} & \makecell[c]{ConvNeXt\\ -UperNet \\ \cite{liu2022convnet}
} & \makecell[c]{PGT*\\(Ours)}\\
\midrule
 $\overline{F_d}$ & 0.615 & 0.545 & 0.625 & 0.706 & 0.582 & 0.615 & 0.698 & \textbf{0.728} \\ 
 p-value & 0.001* & 0.000** & 0.000* & 0.027* & 0.000** & 0.000* & 0.012* & -\\ 
\Xhline{2pt}
\end{tabular}
}
\end{table}

\textbf{The statistical tests. } As shown in Table~\ref{tab:sst}, We calculate $F_d$ of each testing image as sample data and conduct t-test to obtain p-values on the CoNSeP dataset. The p-values are computed between our method and the others.

\section{Conclusion}
We propose a new prompt-based grouping transformer framework that is fully transformer-based, and can achieve end-to-end nuclei detection and classification. In our framework, a grouping-based classifier groups nucleus features into cluster and category embeddings whose correlations with nuclei are used for identifying cell types. We further propose a novel learning scheme, which shares group embeddings with prompt tokens and extracts features guided by nuclei groups with less tuning costs. 
The results not only suggest that our method can obtain competitive performance on nuclei classification, but also indicate that the proposed prompt learning strategy can enhance the tuning efficiency. 
%
%
%
\bibliographystyle{splncs04}
\bibliography{ref}

\end{document}